%% file: paper0007.tex
\RequirePackage{amsmath}
\documentclass[runningheads]{llncs}
\usepackage{xcolor,graphicx,cite,siunitx, amssymb}
\usepackage{hyperref}       % hyperlinks

\input{notations.tex}

\begin{document}

%\title{Parametric and Free-Form Constraints in \Particle-Based Statistical Shape Analysis Using the Penalty Method}

% \title{Parametric and Free-Form Constraints in \Particle-Based Shape Modeling}

\title{Particle-Based Shape Modeling for Arbitrary Regions-of-Interest}

% \title{Particle-Based Shape Modeling for Anatomies with Complex Surface Topologies}

% \titlerunning{Inequality Constraints in Particle-Based Statistical Shape Analysis}

\author{Hong Xu\inst{1} \and
Alan Morris\inst{1} \and
Shireen Y. Elhabian\inst{1}
}
%index{Xu,Hong}
%index{Morris, Alan}
%index{Elhabian, Shireen}
\authorrunning{Hong Xu, Alan Morris, and Shireen Y. Elhabian}

\institute{Scientific Computing and Imaging Institute, School of Computing, \\University of Utah, Salt Lake City, UT, USA \\
\url{http://www.sci.utah.edu}, \url{https://www.cs.utah.edu} \\
\email{\{hxu,amorris,shireen\}@sci.utah.edu}
}

\maketitle              

\input{abstract.tex}
\input{introduction.tex}

\input{methods.tex}

\input{results.tex}
\input{conclusion.tex}

\bibliographystyle{splncs04}
\bibliography{references}

\end{document}

%% file: notations.tex
\newcommand\particle{particle}
\newcommand\aP{\mathbf{\mathcal{P}}}

\newcommand{\SE}[1]{{\textcolor[rgb]{0,0,1}{SE: #1}}}

\definecolor{cadmiumorange}{rgb}{0.93, 0.53, 0.18}
\definecolor{celadon}{rgb}{0.67, 0.88, 0.69}
\definecolor{darkolivegreen}{rgb}{0.33, 0.42, 0.18}
\definecolor{darklavender}{rgb}{0.45, 0.31, 0.59}

\newcommand{\p}{{\bf p}}

\newcommand{\B}{{\bf B}}

\newcommand{\M}{{\bf M}}
\renewcommand{\P}{{\bf P}}

\renewcommand{\S}{{\bf S}}

\newcommand{\ben}{\begin{enumerate}}
\newcommand{\een}{\end{enumerate}}

\newcommand{\cmt}[1]{}

%% file: abstract.tex
\begin{abstract}
Statistical Shape Modeling (SSM) is a quantitative method for analyzing morphological variations in anatomical structures. These analyses often necessitate building models on targeted anatomical regions of interest to focus on specific morphological features.
We propose an extension to \particle-based shape modeling (PSM), a widely used SSM framework, to allow shape modeling to arbitrary regions of interest. Existing methods to define regions of interest are computationally expensive and have topological limitations. To address these shortcomings, we use mesh fields to define free-form constraints, which allow for delimiting arbitrary regions of interest on shape surfaces. Furthermore, we add a quadratic penalty method to the model optimization to enable computationally efficient enforcement of any combination of cutting-plane and free-form constraints.
%
%We use the quadratic penalty method to enforce constraints, limit model construction to regions of interest,
%We propose an extension to \particle-based shape modeling, a widely used SSM tool characterized by its compact representation and computational efficiency but involves expensive computational preprocessing that must be repeated when slight modifications to the input are made. We propose using a signed mesh vector field to define arbitrary free-form constraints (FFCs) as a compact solution to defining highly-customized regions of interest and an extension to the \particle-optimization method to allow these FFCs to work concurrently with parametric constraints, allowing the construction of \particle-based SSMs on these regions of interest without reprocessing the input.
%
%This computationally efficient method uses the quadratic penalty method for inequality constraints to coalesce the constraint objectives as additional terms in the objective function, which can handle building SSMs 
% This method is effective on multiple types of shape representations (meshes, distance transforms, contours). 
%
We demonstrate the effectiveness of this method on a challenging synthetic dataset and two medical datasets. % where certain anatomical features of interest must be excluded.

% Contributions
% \begin{itemize}
%     \item Use a signed mesh vector field to apply arbitrary constraints to surfaces, regardless of representation (mesh, distance transform, contour, etc), to perform particle based statistical shape modeling.
%     \item We use augmented lagrangian constrained optimization to incorporate different types of constraints (free-form, geometric, cutting plane, etc) into a single optimization objective.
%     \item Application and testing on a few medical datasets where certain anatomical features of interest need to be highlighted or certain features must be excluded.
% \end{itemize}
\end{abstract}

%% file: introduction.tex
\section{Introduction}
\label{sec:intro}

% \SE{suggested intro flow: 
% - SSM - what and why?
% - ROI-based SSM - what and why? -- this will introduce the term constraint and why is this important for biomedical and clinical applications
% - implicit vs explicit shape representations for SSM and we focus on explicit and why
% - approaches for explicit and we focus on PSM, what and why? -- this would introduce the notion of particles as a computationally driven approach for SSM
% - previous attempts to constrain particle distribution uses primitives for constraints but arbitrary regions of interest wouldn't admit to these primitives, limiting the topologies to be modeled, also there are technical challenges- scalability for virtual particles
% - in this paper we propose ... we reformulated the PSM cost function within a constrained optimization theme}

% SSM explanation
Statistical Shape Modeling (SSM) is a widespread method used to analyze shape variation across 3D anatomical samples within a population. These analyses are crucial in detecting common morphological pathologies and advancing the understanding of different diseases by studying the form-function relationships between anatomies \cite{atkins2017quantitative, bhalodia2020quantifying, harris2013cam, LenzTalocruralJoint, VANBUURENHipOsteoarthritis, cates2014computational,merle2014many,merle2019high,carriere2014apathy,bruse2016statistical}.
% Constraints motivation paragraph
While building SSMs, certain biomedical and clinical applications require a focus on specific anatomical regions of interest (ROIs) to tailor the analysis to precise morphological features (e.g. \cite{audenaert2019sexualDimorphism, LenzTalocruralJoint, RTW:Dat2009, jacxsens2019coracoacromial, jacxsens2020thinking, atkins2017quantitative, atkins2019two, atkins2022prediction}). Such applications might require excluding certain surface aspects, modeling certain regions in isolation, or a mix of these. ROI definition without altering the input shape has been achieved using \textit{constraints}, mathematical delimiters that restrict model construction to certain surface areas \cite{RTW:Dat2009}. 
%This method is applied to Point Distribution Model (PDM) method for constructing  SSMs, where surface statistics are represented by populating shape surfaces with points/particles/landmarks that maintain relative spatial correspondence across all shapes in the population. 
Our approach focuses on redesigning the constraint application method to improve its functionality, flexibility, and efficiency during SSM construction.

% Landmarks vs deformation fields
To construct such SSMs, two distinct families of shape representations can be used to allow for statistical analysis, \textit{deformation fields} and \textit{landmarks}. Whereas the former encodes \textit{implicit} transformations between cohort samples and a pre-defined (or learned) atlas, the latter uses \textit{explicit} landmark points spread on shape surfaces that correspond across the population \cite{sarkalkan2014statistical, thompson1942growth}. 
% Landmark-based SSM motivation
We focus on the latter approach given its extensive use due to its simplicity, computational efficiency, and interpretability for statistical analysis \cite{sarkalkan2014statistical,zachow2015computational}. 
% Automatic landmark placement
Although landmarks used to be manually placed on specific anatomical features of interest, the modern convention uses dense automatically-placed landmarks obtained through computational methods, such as minimum description length (MDL) \cite{davies2002minimum}, and particle-based shape modeling (PSM) \cite{cates2017shapeworks, cates2007shape}). % \SE{Would constraints work with MDL or LDDMM?}
% PSM pick motivation
We utilize PSM, an efficient and robust entropy-based optimization method that creates a system of dense landmarks or \textit{\particle s}, which conform to all population shape surfaces while maintaining correspondence across them. 
% Constraints motivation
% However, the efficiency of the PSM method is somewhat overshadowed by an arduous preprocessing of the input images, which requires anatomical and technical expertise involving anatomy segmentation and a series of transformations. This type of inefficiency is exacerbated when a specific focus on partial regions of interest is required, in which case, certain preprocessing steps must be repeated to create more targeted SSMs. These targeted models are imperative for building certain shape models where certain areas must be modeled in isolation, must be excluded, or anatomy boundaries are ambiguous (e.g. \cite{audenaert2019sexualDimorphism, LenzTalocruralJoint, RTW:Dat2009}). 
%\SE{here we need several biomedical/clinical applications with references that can benefit from constrains.}

% Virtual particles and method summary
A previous attempt to constrain PSM particle distributions uses geometric primitives in the form of spheres or cutting planes to exclude regions \cite{RTW:Dat2009}. This exclusion is achieved by projecting \textit{virtual \particle s} onto these geometric primitives (represented as parametric constraints), relying on the entropy objective to repel landmark \particle s away from these areas. Such an approach has the advantage of not altering input surfaces, which can otherwise distort morphology or necessitate manual expert-driven reprocessing of data. However, it falters when arbitrary regions of interest cannot be expressed via geometric primitives, limiting the topologies to be modeled. It also exhibits poor scaling due to it requiring an additional set of projected virtual \particle s per constraint. Thereby, to address these shortcomings in the existing literature, we propose the use of the quadratic penalty method in the optimization to allow the simultaneous and scalable application of cutting-plane, spheres, other primitive constraints, as well as a proposed method of defining arbitrary surface constraints, or \textit{free-form constraints} (FFCs). This method provides both flexibility in the definition of constraints to define ROIs and scalability with large-scale or heavily constrained populations without the need to reprocess data.

%% file: methods.tex
\section{Method}
\label{sec:method}

% PSM problem statement
The aforementioned automatic landmark placement methods take in a population of $I-$ shapes $\mathcal{S} = \{\S_i\}_{i=1}^I$ (binary segmentations, meshes, or n-dimensional contours), and obtain \particle s $\aP = \{\P_i\}_{i=1}^I$ %by optimizing an objective $f(\aP)$
where the $i-$th shape point distribution model (PDM) is denoted by $J-$particles $\P_i = [\p_{i,1}, \p_{i,2}, \cdots, \p_{i,J}]$, where $\p_{i,j} \in \mathbb{R}^3$. 
Such \particle s are obtained by optimizing an objective $f(\aP)$, which give

\begin{equation} \label{eq:sws}
\begin{split}
    f(\aP) = H(\aP) - \sum_{j=1}^J H(\P_i),
\end{split}
\end{equation}
where $H$ is an estimation of the differential entropy. 
The \particle s enable quantifying subtle differences and computing shape statistics (e.g., by performing the principal component analysis (PCA) on corresponding \particle s) by providing a population-specific  anatomical mapping across the given cohort.

%Constraint problem statement
We constrain each shape $\S_i$ by $M_i-$inequality constraints in the form $g_{i,m}(\p) \leq 0$, where $g_{i,m}(\p)$ is a differentiable function. These parametric constraints can be in the form of cutting planes or spheres as showcased in \cite{RTW:Dat2009} (by using the equations of planes or spheres), other parametric delimiters, or free-form constraints, which allow arbitrary surface region definition. These constraints limit the distribution of \particle s to regions that satisfy the inequality, a region more easily demarcated using parametric constraints in some anatomies, and/or free-form \textit{surface-painting} in others.

% Section overview
In this section, we describe the use of a quadratic penalty method to allow efficient and simultaneous enforcement of an arbitrary number of parametric constraints, and the use of signed mesh vector fields to build free-form constraints that allow arbitrary surface region isolation. We will also showcase a friendly graphical interface to define these constraints.

%%%%%%%%%%%%%%%%%%%%%%%%
% Quadratic Penalty
%%%%%%%%%%%%%%%%%%%%%%%%

\subsection{Quadratic Penalty for Efficient Constrained PDM Construction}
\label{subsec:quadPen}
% Explain objective
We define an extended objective function to express this constrained optimization problem in an unconstrained form. For each constraint function in the form $g_{i,m}(\p) \leq 0$, we add a quadratic penalty term $g_{i,m}^+(\p) = \max(0, g_{i,m}(\p))$ to the optimization objective, yielding

\begin{equation} \label{eq:rbf}
\begin{split}
    F(\aP) = f(\aP) + \sum_{i=1}^I \sum_{m=1}^{M_i} \sum_{j=1}^J g_{i,m}^+(\p_{i,j}).
\end{split}
\end{equation}

We optimize this objective function using a Gauss-Seidel gradient descent scheme, with the second term preventing \particle s from violating constraints, hence restricting their movement exclusively to feasible regions. This method scales linearly with respect to the number of particles per shape, whereas the virtual particle model \cite{RTW:Dat2009} scales quadratically.

%%%%%%%%%%%%%%%%%%%%%%%%
% Free-Form Constraints
%%%%%%%%%%%%%%%%%%%%%%%%

\subsection{Free-Form Constraints}

\begin{figure}
\includegraphics[width=\textwidth]{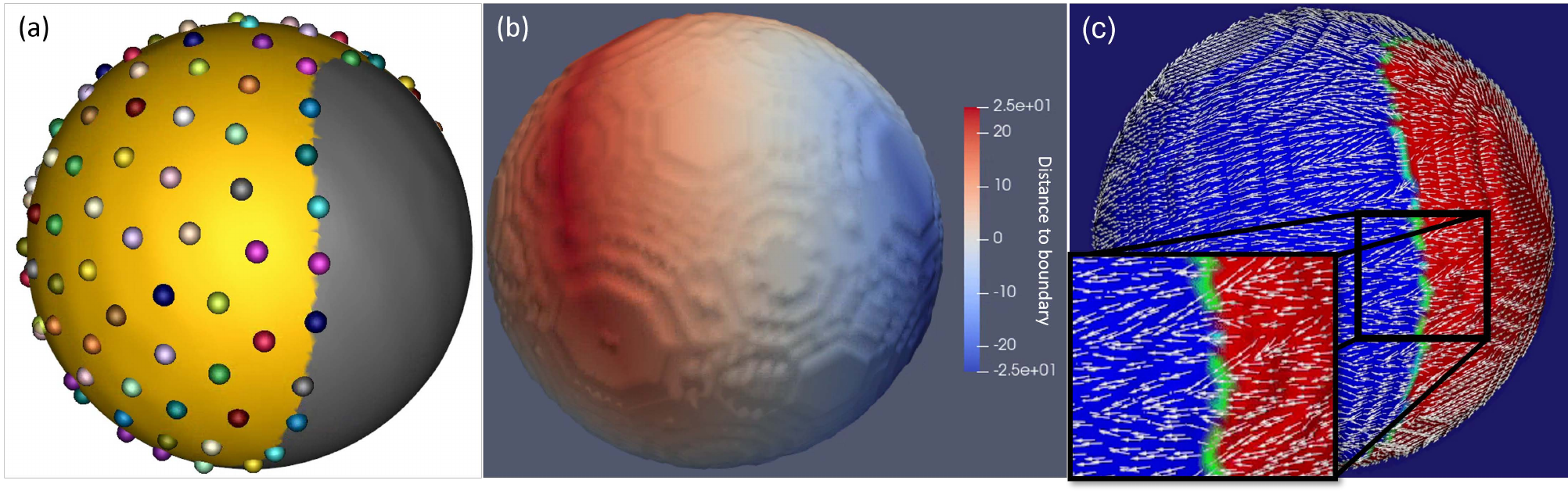}
\caption{(a) Constrained \particle\ distribution on a sphere, where yellow illustrates the feasible region of the constrained area where particles are allowed to be distributed, the gray is the infeasible region where if particles were to be there, they would be violating the constraint. (b) Distance field $\M^d_i(\p)$ of signed geodesic distances to the surface of every mesh vertex. (c) Gradient field $\M^g_i(\p)$ on the mesh surface at every mesh vertex represented using white arrows and the blue surface as the feasible region.
}\label{demonstration}
\end{figure}

We express free-form constraints in the same form $g_{i,m}(\p) \leq 0$ for each shape $\S_i$ by attributing a distance and gradient field onto each vertex of a mesh $\M_i$. Any feasible region on the surface of $\M_i$ can be delineated by a set of surface boundaries $\mathcal{B}_i =[\B_{i,1}, \B_{i,2}, \cdots, \B_{i,B}]$, which are represented as vertex loops on the mesh surface. A distance field query for a particle $\p$, denoted $\M^d_i(\p)$, provides the signed geodesic distance to the closest constraint boundary $\B_{i,*}$ from the projection of $\p$ onto $M_i$, illustrated in Figure \ref{demonstration} (b). Similarly, a gradient field query, denoted $\M^g_i(\p)$, would provide the gradient direction, shown in Figure \ref{demonstration} (c). Ultimately, the mesh $\M_i$ together with its fields $\M^d_i$ and $\M^g_i$, can approximate the distance and first-order gradients over near-surface points, effectively simulating a differentiable function $g_{i,m}(\p)$. When integrated into the aforementioned penalty method \ref{subsec:quadPen}, this approach can enforce arbitrary surface constraints.

%%%%%%%%%%%%%%%%%%%%%%%%
% Graphical interface tool
%%%%%%%%%%%%%%%%%%%%%%%%

\subsection{Graphical Interface Tool}

\begin{figure}
\includegraphics[width=\textwidth]{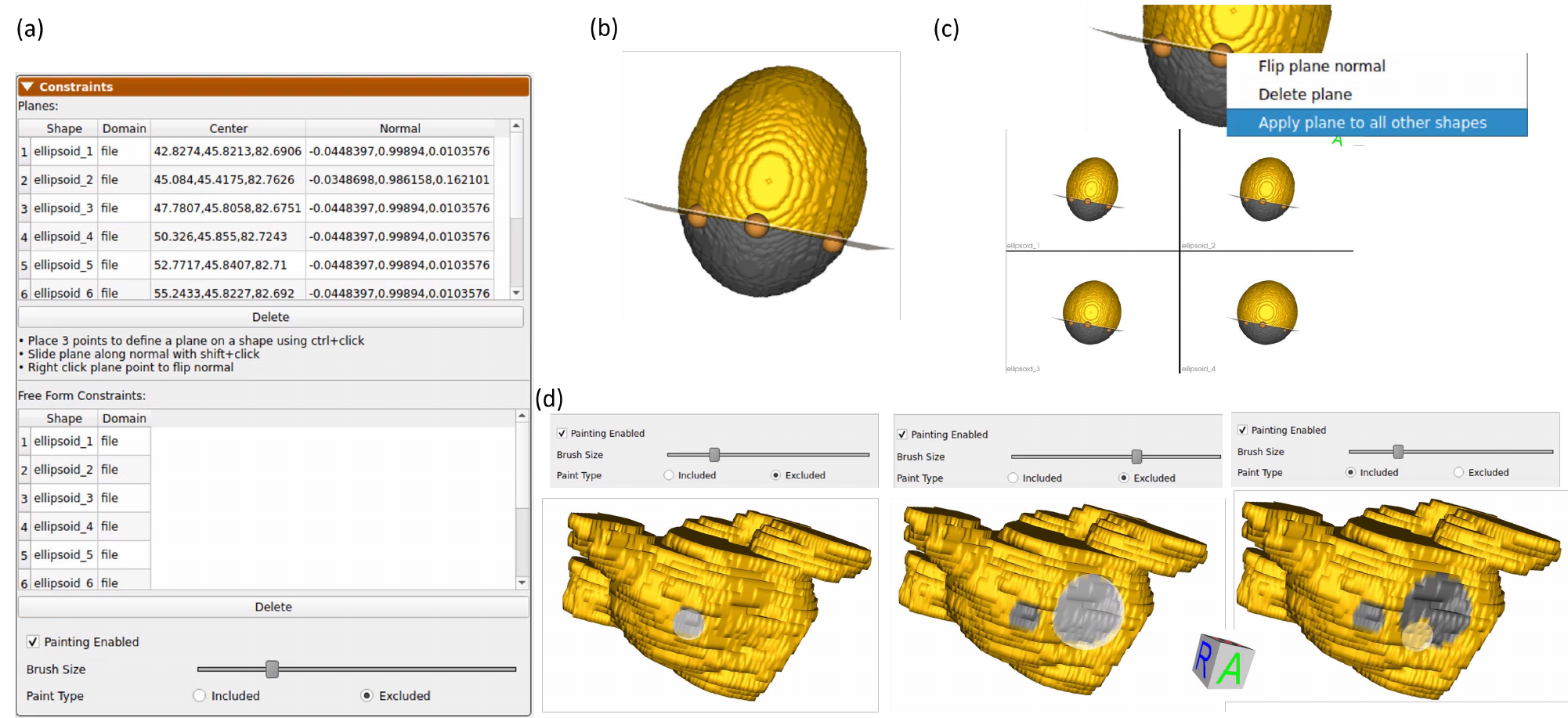}
\caption{(a) The constraint panel shows the constraints that have been defined and the tools to define the constraints. (b) Cutting-plane constraints are defined by ctrl-clicking 3 points on the shape surface. (c) Constraints can be flipped or applied to all other shapes via the right-click menu. (d) FFCs are defined with a painting tool with different brush sizes and options to customize included and excluded areas. We show how the painting of excluded areas of different sizes applied to a segmentation of a left atrium.
}\label{tool_showcase}
\end{figure}

We include a graphical interface tool that can define cutting planes and FFCs and can roughly propagate these to all shapes in the population. Cutting planes are defined by prompting 3 points that the user can pick that are on the shape surface, and can be copy-pasted into all other shapes. FFCs are defined using a "painting" tool that can define included and excluded areas with an adjustable brush size. This tool allows precise and arbitrarily customizable definition of constraints. An FFC on a single shape can be propagated to others using deformation parameters computed from image registration. This functionality is also included. All the graphical interface functionality is illustrated in figure \ref{tool_showcase}.

%% file: results.tex
\section{Results}
\begin{figure}
\includegraphics[width=\textwidth]{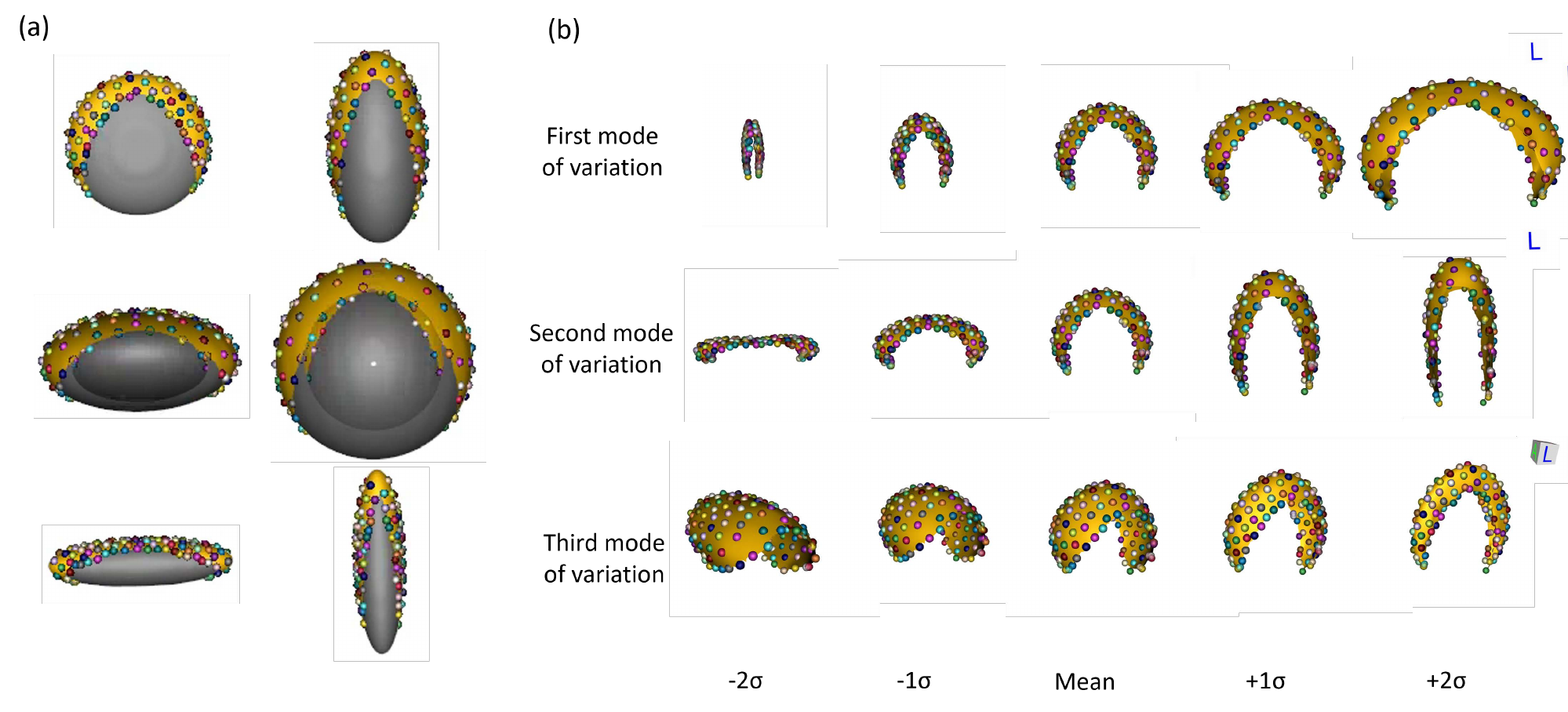}
\caption{(a) Sample ellipsoids from the dataset with feasible regions in yellow and restricted regions in grey. (b) The first three modes of variation in the dataset, which show the variation in corresponding major axes.
}\label{ellipsoid}
\end{figure}

% Intro and ellipsoid
We demonstrate our results by integrating our method into an open-source implementation of the particle-based shape modeling (PSM) framework, ShapeWorks \cite{cates2017shapeworks}, and produce SSMs from three datasets. The first is a synthetic dataset of ellipsoids that vary between values of 10, 20, 30, and 40, in each of their three major axes, totaling 64 ellipsoids. These ellipsoids are constrained by a free-form boundary that divides each ellipsoid into upper and lower halves by a full period of a sine wave projected onto the surface, providing a challenging but uniformly delimited population of shapes. Figure \ref{ellipsoid} shows a few examples and the modes of variation from the SSM. The constraints have the desired effect, and the modes of variation meet expectations as they mimic the variation in the three major axes.

\begin{figure}
\includegraphics[width=\textwidth]{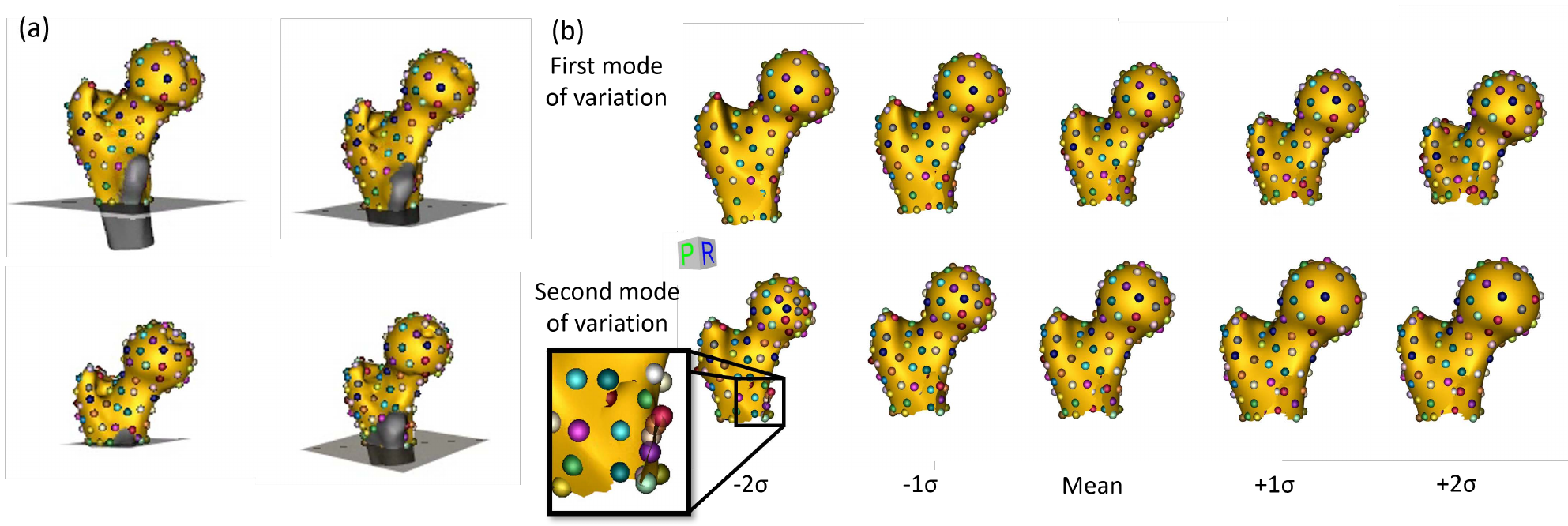}
\caption{(a) Example of defined constraints. The feasible region is shown in yellow and the constrained region in grey. (b) The first two modes of variation in the dataset. Notice that \particle s are excluded from the lesser trochanter.
}\label{femur}
\end{figure}

% Femur dataset
The second is a dataset of 25 computerized tomography (CT) femurs, where the region of interest is the proximal femur sans the lesser trochanter (femoral head, neck, and greater trochanter). For each shape, we use a cutting plane constraint to exclude the shaft and a free-form constraint to exclude the lesser trochanter. Figure \ref{femur} illustrates a few examples and the first two modes of variation. We observe that a cutting plane allows a more straightforward exclusion of the shaft whilst the FFC precisely excludes the lesser trochanter. The constraints restrict the movement of particles to the feasible region as expected, and the modes of variation meet expectation as well.

% Left Atrium dataset
%The third dataset comprises 21 segmentations of left atria models obtained from MRIs. \TODO{Explain why the area must be isolated.} Therefore, we exclude the \SE{region} by painting the excluded region. Figure \ref{la} showcases some examples of the shape and the first three modes of variation. The models meet expectations.

\begin{figure}
\includegraphics[width=\textwidth]{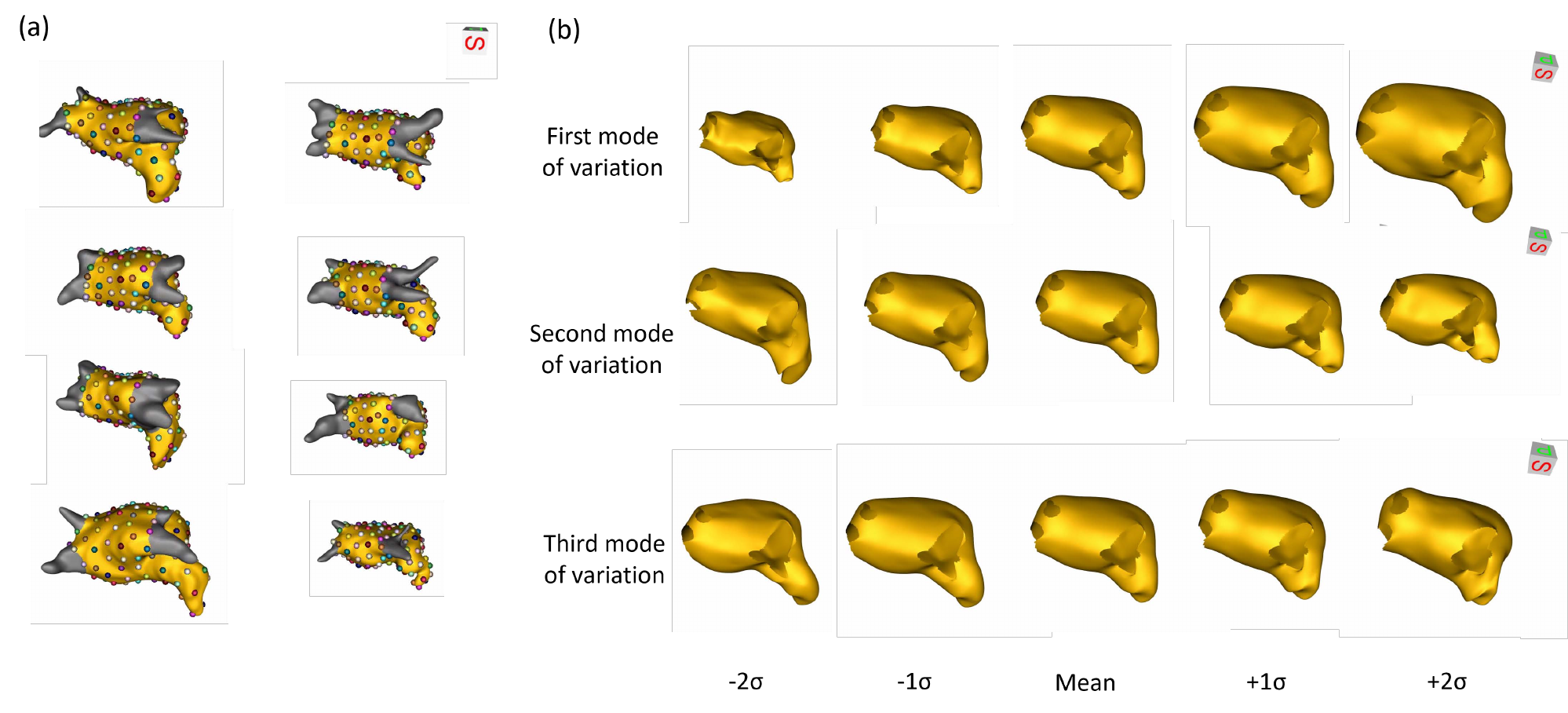}
\caption{(a) Example of defined constraints where the left atrium (in yellow) segmentations have the pulmonary veins excluded (in grey). (b) The first three modes of variation in the dataset. Notice how the pulmonary vein areas remain hollow.
}\label{la}
\end{figure}

The third dataset comprises 21 segmentation of left atria models obtained from MRIs.  The pulmonary veins represent the area of greatest variation both in anatomical structure (e.g. number of veins, common veins, etc) as well as greatest variability in segmentation by expert observers (e.g. length into vein to segment).  While the position of veins may be important from a shape modeling perspective, their exact shape is not particularly relevant to LA shape morphology.  Thus, we paint a free-form constraint exclusion area around the veins.  Figure \ref{la} showcases some examples of the shape and the first three modes of variation.  The models meet expectations.

%% file: conclusion.tex
\section{Conclusion}
\label{sec:conclusion}

We demonstrate a flexible and more scalable approach to define regions of interest in fully-groomed shapes for landmark-based statistical shape modeling by allowing arbitrary definition of surface constraints via FFCs and incorporating mixed constraint types into the optimization. This significantly improves the usability of PSM methods, obviating the need for reprocessing datasets. Future work includes the automatic propagation of constraints to the entire cohort given manual definitions on certain representative shapes.